\documentclass[10pt,conference]{IEEEtran}
\IEEEoverridecommandlockouts
\PassOptionsToPackage{hyphens}{url}\usepackage{hyperref}
\usepackage{cite}
\usepackage{amsmath,amssymb,amsfonts}
\usepackage{graphicx}
\usepackage{mdframed}
\usepackage{multirow}
\usepackage{textcomp}
\usepackage{xcolor}
\usepackage{url}
    
\begin{document}

\newtheorem{hypothesis}{Hypothesis}

\definecolor{dkgreen}{RGB}{0, 128, 0}

\title{Leveraging LLMs for Legacy Code Modernization: Challenges and Opportunities for LLM-Generated Documentation}

\author{\IEEEauthorblockN{Colin Diggs, Michael Doyle, Amit Madan, Siggy Scott, \\Emily Escamilla, Jacob Zimmer, Naveed Nekoo, \\Paul Ursino, Michael Bartholf, Zachary Robin, Anand Patel, \\Chris Glasz, William Macke, Paul Kirk, Jasper Phillips, \\Arun Sridharan, Doug Wendt, Scott Rosen, \\Nitin Naik, Justin F. Brunelle, Samruddhi Thaker\textsuperscript{\textsection}}
\IEEEauthorblockA{\textit{The MITRE Corporation} \\
McLean, VA\\
cdiggs@mitre.org}
}

\maketitle

\begingroup\renewcommand\thefootnote{\textsection}
\footnotetext{Due to the size of the team, we list everyone here but see the first author's contact information below as the corresponding author}
\endgroup

\begin{abstract}
    Legacy software systems, written in outdated languages like MUMPS and mainframe assembly, pose challenges in efficiency, maintenance, staffing, and security. While LLMs offer promise for modernizing these systems, their ability to understand legacy languages is largely unknown. This paper investigates the utilization of LLMs to generate documentation for legacy code using two datasets: an electronic health records (EHR) system in MUMPS and open-source applications in IBM mainframe Assembly Language Code (ALC). We propose a prompting strategy for generating line-wise code comments and a rubric to evaluate their completeness, readability, usefulness, and hallucination. Our study assesses the correlation between human evaluations and automated metrics, such as code complexity and reference-based metrics. We find that LLM-generated comments for MUMPS and ALC are generally hallucination-free, complete, readable, and useful compared to ground-truth comments, though ALC poses challenges. However, no automated metrics strongly correlate with comment quality to predict or measure LLM performance. Our findings highlight the limitations of current automated measures and the need for better evaluation metrics for LLM-generated documentation in legacy systems.
\end{abstract}


\section{Introduction}

Legacy systems are systems that contain outdated computer software. These systems are often written in antiquated languages like COBOL, MUMPS, or historical dialects of mainframe assembly which make maintenance and further development challenging. In the U.S. federal government, a large fraction of legacy systems range in age from 30 to more than 60 years old \cite{gao2019}, posing significant efficiency, maintenance, staffing, and security challenges. Modernization of legacy systems aims to improve maintainability of development of these critical systems. However, the work of modernizing legacy software systems has become well-known as a ``wicked problem" that has largely defied decades of well-meaning initiatives and policies \cite{alexandrova2015legacy,eganItif}.   

With the advent of large language model (LLM) approaches to code synthesis and translation, there is widespread optimism that artificial intelligence is poised to increase efficiency and reduce the risks of software modernization \cite{DiegoLoGiudice2024,Aulbach2024}. However, it is not yet clear whether the code-understanding abilities of LLMs are mature enough to have a significant impact on the problem \cite{chen2021evaluating,greengard2023ai,fan2023large,kc2023neural}. Many shortcomings of LLMs have been demonstrated on software understanding---which include tasks like documentation generation, code translation, and code synthesis---even on clean code and mainstream languages \cite{jesse2023large,kabir2024stack,vaithilingam2022expectation,weisz2022better}. Developers and decision makers are justifiably uneasy about adopting these tools into modernization workflows for complex, real-world legacy systems---many of which fill safety-critical roles where the introduction of inadvertent bugs or functionality changes can raise significant risks.

Studies on LLM-based direct code translation reveal limitations even for modern languages such as C, Java, and Python \cite{pietrini2024bridging}. There is a gap in the literature for similar studies with legacy languages \cite{pietrini2024bridging}. As such, direct translations by LLMs cannot be trusted for the modernization of critical legacy systems with no tolerance for error. However, in the emerging landscape of LLM-based code understanding tasks, documentation generation, or code summarization, emerges as a compelling modernization strategy. Even in cases where LLM-driven code \emph{translation} might not be trustworthy, automatically generated documentation may accelerate code understanding to facilitate translation, refactoring, and maintenance if it can be generated consistently and accurately. While documentation generation methods have advanced greatly in recent years \cite{leclair2019neural,leclair2020improved,wan2018improving}, they have been trained and evaluated primarily on mainstream languages like C, Python, and Java, and on relatively short, simple programs with limited complexity \cite{leclair2019recommendations}. This makes it difficult to infer whether results on mainstream benchmarks will generalize in useful ways to legacy software---which is often written in niche or antiquated languages while also exhibiting extreme complexity.

Here we study the performance of an LLM-based approach at generating documentation for legacy languages, focusing on two real-world legacy datasets: an electronic health records (EHR) system written in the MUMPS\footnote{Massachusetts General Hospital Utility Multi-Programming System---a high-level language and database developed in the 1960s for managing patient medical records \cite{sherertz1980evolution}} language, and a set of open-source applications in IBM mainframe assembly language code (ALC).\footnote{In the IBM mainframe industry, assembly code is routinely referred to by the longer name ``assembly language code."} Using LLMs to generate documentation raises two immediate challenges.  First, no standard prompting strategy has emerged in the community for reliably generating line-wise code comments.  Our preliminary efforts, for example, found that direct LLM prompting often generates incomplete output, and may perform undesired operations such as modifying the code before generating comments for it. Second, no widely accepted evaluation metrics exist for measuring the quality of documentation of code via comments.  Evaluating generated comments is fundamentally important for industrial applications, where the quality of the output determines whether an automated tool is ready to be used on production codebases.  Manual human evaluation of large numbers of comments is costly, but evaluation must be performed on each new system (even on mainstream languages, the quality and reliability of generated documentation can vary dramatically across codebases \cite{leclair2019recommendations,stapleton2020human}).  There is thus a need for automated metrics of documentation quality that can be used to test and develop documentation generation tools.  

In this work, we address these problems by testing comment-generation and evaluation strategies, with the objective of providing baseline results to understand the ability of mainstream LLMs to assist in documenting legacy code:
\begin{itemize}
    \item Our \textbf{prompting strategy} for generating line-wise code comments with LLMs avoids the problems of incomplete comment generation and undesired code modification.  We evaluate its performance in terms of a novel \textbf{scoring rubric} that measures human perceptions of completeness, readability, usefulness, and hallucination---developed in close collaboration with subject-matter experts (SMEs) with many years of legacy code experience.  
    \item We then measure the efficacy of three classes of metrics that can be automatically computed at scale without manual human evaluation to see how well they predict or measure (human-rated) documentation quality:
    \begin{itemize}
         \item \textbf{static code complexity} metrics (e.g., cyclomatic and Halstead complexity),
         \item \textbf{runtime metrics} on the LLMs themselves (e.g., processing time), and 
         \item mainstream \textbf{reference-based metrics} (e.g., BLEU and ROUGE) that make use of human-written ground truth comments.
    \end{itemize}
\end{itemize}
Ultimately, we hope this line of inquiry can provide organizations with a framework that can be used to decide whether a given model will perform well enough on their unique codebases to provide value through (partial) automation of software modernization efforts.

\section{Background}

Legacy software poses unique economic and technical challenges. Here we discuss the problem of software modernization and the prospects of LLM-based solutions. We give special attention to machine-assisted documentation generation and the lack of effective methods for evaluating the quality of generated documentation.

\subsection{Legacy Software}


Government agencies need to migrate their obsolete IT to less expensive, more agile systems that mitigate security risk, and provide more efficient and impactful service delivery for taxpayers. The U.S. federal government has been trying to overhaul its IT infrastructure for decades. Agencies are falling short in their IT modernization initiatives, spending almost 80\% of their IT budgets on operations and maintenance, compared with around half in the private sector \cite{eganItif}. According to the Government Accountability Office (GAO), eight of the ten critical federal IT legacy systems that were most in need of modernization did not have documented plans for modernizing their systems or had incomplete plans \cite{gao2023}.

These legacy systems support important missions like wartime readiness and the operation of dams and power plants. They also host sensitive taxpayer and student data. The GAO has reported on these systems since 2016, highlighting security risks, unmet mission needs, and the increased maintenance costs associated with outdated systems. Most recently, GAO reported that the Internal Revenue Service (IRS) has systems that are more than 60 years old, with some operating software that is up to 15 versions out of date \cite{gao2016, gao2019, gao2023-2}. The recent Federal Aviation Administration (FAA) systems outage that canceled 1,300 flights and delayed more than 10,000 in a single day\footnote{CNN, ``FAA is years away from upgrading the system that grounded all US flights," January 3rd, 2023---\url{https://www.cnn.com/2023/01/12/tech/faa-notam-system-outage/index.html}.} highlights both the criticality of these legacy systems and the impact that a single outage can have on our transportation network and on the daily lives of thousands of individuals.

Traditional approaches to converting legacy code into validated, modernized analogues often involve complex rules and pattern-matching approaches that are costly to develop \cite{NEWCOMB2010301}. As Zaystev et al.\ observe, even the creation of parsers, compilers, and other tools needed to begin a legacy conversion project can take years \cite{zaytsev2017open}. While some traditional rule-based and symbolic conversion tools exist and offer formal guarantees that their output is functionally correct, these tools can typically only be applied to very narrow cases and domain-specific applications \cite{bhatia2024can}.

A study by Pietrini et al.\ details the complexities of modernizing a critical system from legacy and modern Fortran code to Python \cite{pietrini2024bridging}. They identified that manual translation of legacy code is a time-consuming and error-prone process and that there is a conspicuous gap in both the research and tools available for the automated translation of legacy languages to modern languages. To address this gap, Pietrini et al.\ leveraged GPT-4 for direct translation of the Fortran code into Python. One of the noteable challenges identified by Pietrini et al.\ was the lack of documentation in the legacy code which made it challenging to determine the validity of the translated code. The importance of documentation in the modernization of legacy systems motivates our experiment to leverage LLMs to generate documentation for legacy systems.

\subsection{LLM-Based Code Understanding}

But even on mainstream languages, LLM-based code translation and synthesis models show significant limitations upon closer inspection.
On many code-related tasks, LLMs introduce simple bugs and incorrect information at high rates \cite{jesse2023large}. A recent evaluation by Pan et al. on 1,700 different open-source code examples, for instance, found that LLM translation among C, C++, Go, Java, and Python could produce code that passed unit tests at most 47.3\% of the time and only 2.1\% in the worst-performing model \cite{pan2024lost}. Based on this data, they propose a taxonomy of 15 categories of translation bugs that LLMs are prone to. Additionally, as Kabir et al. show in their study of StackOverflow answers, human reviewers may fail to detect a large fraction of these errors \cite{kabir2024stack}. In practical (human-machine teaming) terms, efforts to study the impact of LLM-based systems on developer productivity have yet to reliably demonstrate beneficial effects \cite{vaithilingam2022expectation,weisz2022better}. Wu et al., meanwhile, show that code synthesis performance plummets when tasks deviate from historical training data \cite{wu2023reasoning}---an especially worrying observation in the context of legacy systems, which often involve rare languages and unusual coding patterns that are not reflected in public training data. 

There is a significant literature gap around the evaluation of LLM-based code understanding for legacy languages, as research in this area is still at an early stage.  Perhaps with the exception of neural decompilation (i.e., converting assembly into higher-level languages) \cite{liang2021neutron,fu2019coda,katz2019towards}, few benchmarks or quantitative results are available for legacy languages. Results that do exist are often reported on open-source datasets which tend to involve much shorter and simpler programs than real-world legacy code.

\subsection{Documentation Generation}

Ensuring the validity of the modernization of legacy systems requires robust understanding of the existing code \cite{pietrini2024bridging}. In the absence of SMEs, proper documentation provides insight into existing functionality. As such, creating documentation is an important subtask of software modernization. Legacy codebases often lack accurate and sufficient documentation to support the efficient creation of downstream models, onboarding materials, requirements, and translated or refactored code. 

While using LLMs to directly translate or generate code into modern languages suffers from the limitations we have described above, there is reason to believe that the code-understanding abilities of foundational models may still be useful to developers as they conduct complex code modernization projects. Documentation generation is one such task where LLM-based tools may be useful in a human-machine teaming setting. Documentation generation involves creating inline comments, higher-level comment blocks, and summaries from source code. Due to the critical nature of legacy systems and low tolerance for error, documentation is necessary to ensure the validity of code translation and other modernization tasks. 

\subsubsection{Template-Based Methods}
Traditional approaches based on manual templates \cite{mcburney2015automatic,moreno2013automatic,sridhara2010towards} and information retrieval \cite{wong2015clocom} can achieve a certain degree of automated generation for high-level languages like Java. But these methods rely heavily on variable names, object-oriented structure, and direct access to huge databases of related code (such as all of GitHub) to recast source code into natural language.

\subsubsection{Neural Methods}
More recently, a variety of neural approaches to documentation generation have been developed that improve on rule- and information retrieval (IR)-based methods---see surveys by Zhang et al.\ \cite{zhang2022survey} and Zhu et al.\ \cite{zhu2019automatic}. Notable examples have been based on long-term short-term networks (LSTMs) \cite{lin2021improving,hu2020deep,iyer2016summarizing}, deep reinforcement learning \cite{wan2018improving}, and graph neural networks \cite{leclair2020improved}. Some methods represent code as tokens, much like natural language---treating documentation generation as a machine-translation problem. Other approaches leverage abstract syntax tree (AST) representations to integrate explicit syntactic information into a model's input \cite{leclair2019neural,hu2020deep,gao2023code}. While several approaches have applied neural attention mechanisms such as transformers to the documentation-generation problem \cite{lin2021improving,iyer2016summarizing}, researchers have generally trained these models from scratch on each new dataset.

\subsubsection{Foundation Models}

The advent of (reusable) foundation models \cite{bommasani2021opportunities} have offered a third major methodological shift in the documentation generation literature. Gu et al.\ provide a preliminary example, in which they create a combined foundation model by using  CodeBERT as an encoder for code snippets and GPT-2 as a decoder to generate comments (where they add a trainable Gaussian noise layer in between to convert embeddings outputted by the former into input for the latter) \cite{gu2022assemble}.

LLMs have been found to produce documentation that is similar to or better than original, developer-created documentation \cite{kahn2022automatic, dvivedi2024comparative}. Codex, a GPT-3 model trained on GitHub data, generated documentation that was similar in quality to developer-created documentation \cite{kahn2022automatic}. Non-specialized LLMs (GPT-3, GPT-4, Bard, LLama 2) outperformed original documentation for both inline and function-level documentation \cite{dvivedi2024comparative}. In modernization efforts, access to experts in both the legacy and the target languages may be limited, so documentation must be applicable to non-experts. Students found that function-level code explanations generated by an LLM were more understandable and accurate than explanations created by their peers \cite{leinonen2023comparing}. Thus, LLMs are capable of creating code documentation that benefits non-experts.

However, it is notable that performance varies across languages \cite{kahn2022automatic}. Some proposals for LLM usage focus on using AI-assisted UIs to generate code explanations on demand, rather than integrating documentation directly into the codebase---thereby possibly reducing the negative impact that hallucinations may incur \cite{nam2024using}. Regardless, existing studies assess LLM performance on simple functions and modern lanugages (e.g. JavaScript, Java, PHP, Python) \cite{kahn2022automatic, leinonen2023comparing, dvivedi2024comparative, nam2024using}. This study addresses a research gap in documentation generation for real-world complex code written in legacy languages (e.g. MUMPS and ALC).

\subsection{Evaluating Documentation}

As with other code understanding tasks, neural and LLM models for documentation generation continue to exhibit limitations. For example, while they eliminate the narrow reliance that earlier template-based approaches have on specific keywords and variable names, it has been shown that neural models still rely heavily on textual cues in source code to generate documentation \cite{sontakke2022code}.\footnote{Though see Macke \& Doyle for evidence that the presence or absence of comments in the input to code generation systems has little impact on the output \cite{macke-doyle-2024-testing}.} But evaluating documentation quality is a difficult and multi-dimensional problem, and \emph{automating} such evaluations so they can be scaled up to large datasets has proved especially difficult \cite{tang2023evaluating}.  Researchers currently rely primarily on manual human evaluation, typically using ad-hoc rubrics that vary from author to author \cite{hu2022correlating,shi2022evaluation,stapleton2020human}, sometimes in combination with reference-based metrics such as ROUGE-N or BLEU.  The latter have been shown to correlate poorly with human evaluations, however \cite{hu2022correlating,shi2022evaluation,stapleton2020human}.  Tang et al.\ \cite{tang2023evaluating} review some 32 dimensions of documentation quality that have been used in human evaluations, some of which can be integrated into automated approaches for specific kinds of evaluation (such as mapping code examples to associated documentation passages).  Overall, however, there is no well-established methodology for evaluating documentation quality.

\section{Methods}

At a high level, we applied four LLMs to our datasets to generate comments for each line of code, and then asked human experts to review both the generated comments and the original comments in the codebase for quality.  Our first methodological focus is on the comment-generation strategy for ALC and MUMPS: here we will walk through the chunking strategy, prompting method, and parameters used in our experiments, as well as the datasets the themselves.  The next focus area is the evaluation approach---where we will describe the human-review rubric and the automated metrics that we tested for their ability to predict comment quality.

\subsection{Models and Generation Method}
\label{sec:models}

We implemented our comment-generation methods using the open-source Janus framework for LLM-based code modernization.\footnote{\url{https://github.com/janus-llm/janus-llm}}  We focused our experiments on \emph{line-wise comments}---that is, comments that are associated with a specific line of code (appearing either as a block comment immediately preceding the line, or as an inline comment at the end of the line).  Our focus on line-wise comments is motivated by the experience of SMEs with many years of experience maintaining legacy systems implemented on IBM Z-series mainframes.  These experts report to us that the obscurity of low-level logic in legacy software hinders the ability to understand code, model it for refactoring efforts, and to train and onboard new developers.  Better line-wise comments could mitigate this problem.  Broadly, our approach begins by chunking code into syntactically unbroken segments that are small enough to fit into each LLM's context window.  Then we use a prompting strategy that replaces ground-truth (human-written) comments with placeholder identifiers, and asks an LLM to produce comments for each placeholder in a JSON-structured output format.  This strategy prevents the LLM from modifying or hallucinating new code while it adds documentation.

\subsubsection{Chunking strategy}


The choice of method for chunking, or partitioning a source file into pieces to feed into an LLM's context window can in some cases significantly impact the quality of the output. In general, the problem of determining what portion of a codebase should be included to maximize the quality of output is a potentially complex document segmentation or information retrieval problem.  Here, we adopt a segmentation strategy to pack as much code as possible into the context window while limiting the division of logical structures. To achieve this, we adopt a greedy chunking algorithm to partition each source file. The algorithm begins by making each subroutine into a chunk, and then recursively merges the smallest chunk with its neighbors whenever doing so does not yield a chunk larger than half of the model's context window (reserving half the context window for the generated output). The result is that different models receive smaller or larger inputs, depending on what their context window can support.

\subsubsection{Prompt Template}
In preliminary experimentation, we found that a na\"ive prompting strategy of feeding code into an LLM and asking it to add comments to every line was ineffective, as the models would often alter the code to generate documentation---a task outside the scope of the prompt. This was especially a problem in MUMPS, which idiomatically may exhibit some 5--6 separate logical statements on a single line: the LLM would often rewrite  the code to be several lines of code, generating 5--6 separate comments for new lines where a human wrote only one. Code-rewriting during comment generation makes it difficult to preserve a 1-to-1 mapping between generated code and ground truth, human-provided comments.

\begin{figure}
    \includegraphics[width=\columnwidth]{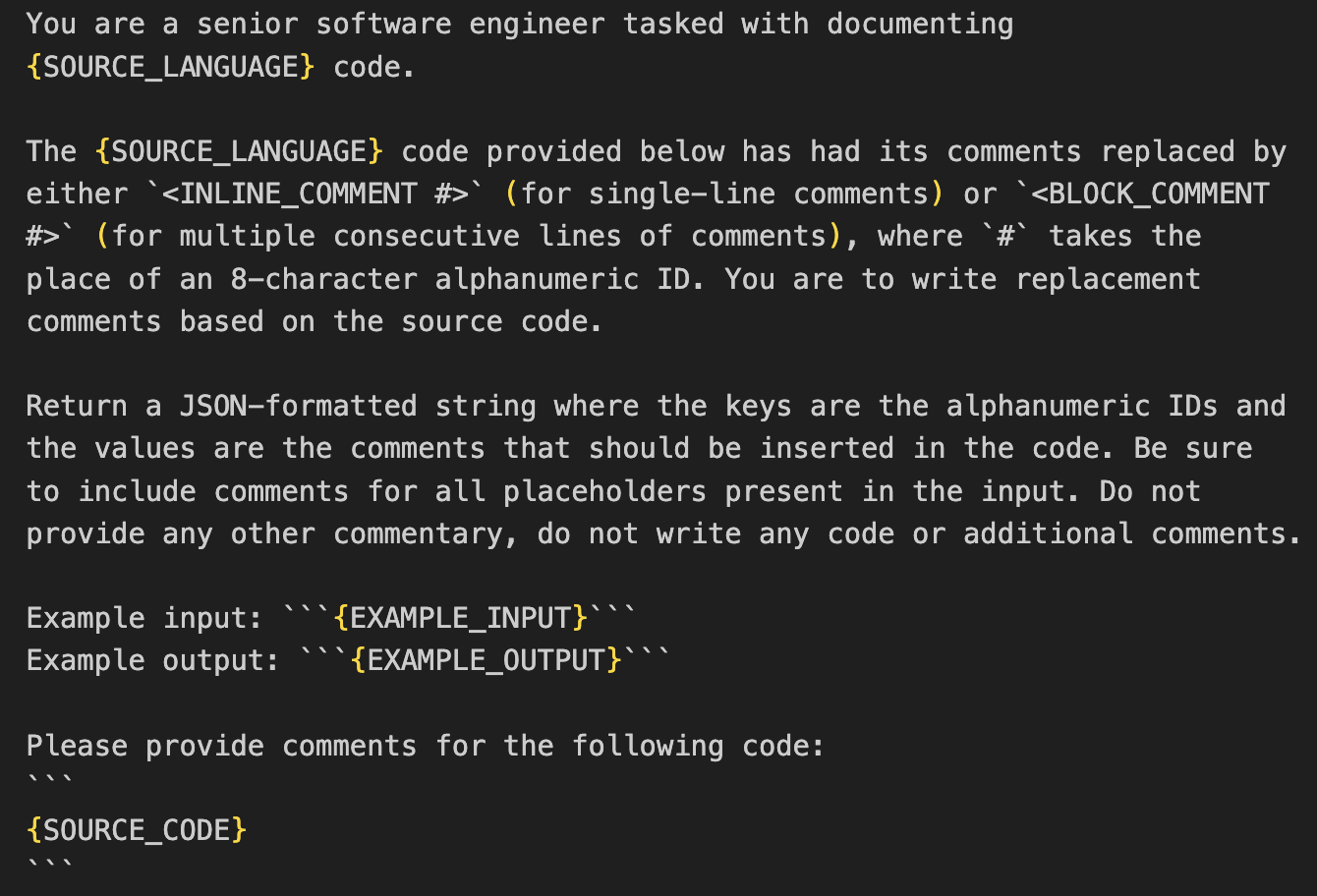}
    \caption{The prompt template we used for line-wise comment generation.}
    \label{fig:prompt}
\end{figure}

\begin{figure}
    \centering
    \includegraphics[width=\columnwidth]{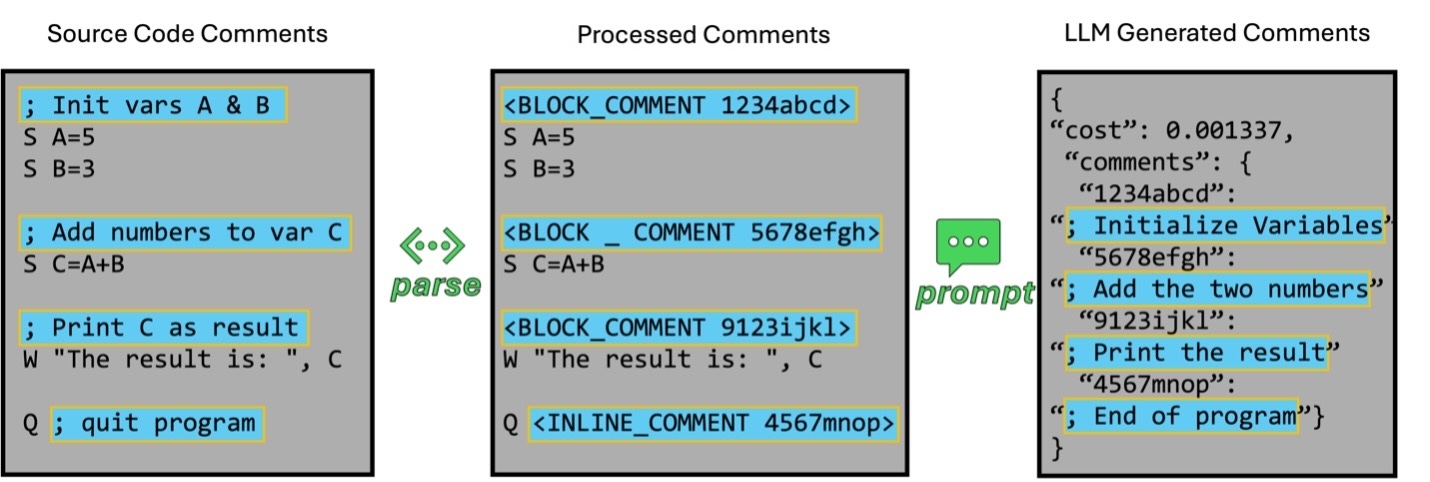}
    \caption{Pre-processing procedure to replace existing in-line comments with unique identifiers.}
    \label{fig:madlibs}
\end{figure}

To resolve this, we devised a novel prompting strategy to discourage the models from altering code:
\begin{itemize}
    \item First we replace all of the human-written (ground truth) comments in each file with placeholder.  The placeholders have the form \texttt{<BLOCK\_COMMENT [id]>} and \texttt{<INLINE\_COMMENT [id]>}, where \texttt{[id]} is a unique alphanumeric identifier.
    \item We then prompted each LLM to generate structured output in JSON format, providing a comment associated with each id.
\end{itemize}
The prompt template is given in Figure \ref{fig:prompt}, and an illustration of the comment replacement process and JSON output is shown in Figure \ref{fig:madlibs}. We found that providing markers allowed the LLM to more accurately identify the location of interest. Using unique, randomly generated identifiers also mitigated hallucination that might occur with an LLM creating its own unique identifiers. Requesting structured output, such as JSON-formatted output, provides two benefits: reduced likelihood of the LLM improvising beyond the input and instructions, and streamlined integration with downstream processes \cite{liu2024we}. Requesting JSON-formatted output also allowed us to easily map the unique identifiers with the comments generated from each of the twelve LLMs.

\subsubsection{Models \& Parameters}

Due to the time required for human evaluation of automated documentation generation, we focus our experiments on four models: Claude 3.0 Sonnet, Llama 3, Mixtral, and GPT-4.  The context windows for these models are shown in Table \ref{tab:models}. The generated comments we obtain from these models were then evaluated for complexity and quality by human evaluators and several automated metrics described below.  We did not perform hyperparameter tuning on the models, but used a temperature value of 0.7 across all models for consistent comparison.

One challenge the models present is incomplete output---sometimes the JSON output from the model fails to produce a comment for all of the placeholders in the input. In this situation, we execute the model multiple times until a comment has been generated for every placeholder. While this approach resulted in additional processing time, it prevented outright failure.  We include retries of this kind when measuring the cost and execution time of the models.

\begin{table}[]
\centering
\begin{tabular}{|l|l|c|}
\hline
Family & Name & Context Window\\ \hline
\multirow{1}{*}{Anthropic} & Claude 3.0 Sonnet & 200k tokens\\
 \hline
\multirow{1}{*}{Meta Llama} & Llama 3 Instruct (70B) & 8k tokens\\
 \hline
\multirow{1}{*}{Mistral} & Mixtral 8*7B & 32k tokens\\
 \hline
\multirow{1}{*}{OpenAI} & GPT-4 Turbo Preview & 128k tokens\\
 \hline
\end{tabular}
\caption{The LLMs used for these experiments.}
\label{tab:models}
\end{table}

\subsection{Datasets}

There is a lack of quality benchmark datasets which makes the development of advanced documentation generation techniques difficult \cite{leclair2019recommendations}. This challenge is compounded for \emph{legacy} software, which is often written in rare or domain-specific languages. Open-source examples of code in these languages, are often relatively small and simple programs, and may lack the enormous complexity and technical debt that often plague real-world legacy systems. In the absence of a benchmark dataset of legacy code, we created two datasets that are representative of real-world legacy systems: one for MUMPS and one for ALC.  We will describe how files from these corpora are chunked into segments to be loaded into LLMs below, in Section~\ref{sec:models}.

\subsubsection{MUMPS Corpus}
We used the WorldVistA MUMPS implementation of VistA for our MUMPS dataset, with a particular focus on the Incomplete Records Tracking module. VistA is the name given to the EHR modules that make up the healthcare information technology system of the Veterans Health Administration. The open-source version of VistA\footnote{\url{https://github.com/WorldVistA/VistA}}, which consists of 175 modules, 36,000+ routines (files), and nearly 400,000 subroutines, is maintained by the Open Source Electronic Health Record Alliance (OSEHRA). From OSEHRA VistA, we selected the Incomplete Record Tracking module as representative of an average module in terms of the number of comments, routines, and subroutines it contains. The Incomplete Record Tracking module consists of 78 files, containing many of the conventions and styles found within older and newer MUMPS code. It is also a complete software module, making it a better candidate for summarization tasks than a random sample of all files within the entire codebase. For this reason, we consider Incomplete Record Tracking a representative module of the VistA codebase for our experimentation. In total, the MUMPS corpus contains 5,107 lines of code and 235 developer comments across 78 files. 

\subsubsection{ALC Corpus}
We used the zFAM repository of the open-source Walmart codebase\footnote{\url{https://github.com/walmartlabs/zFAM}} for our ALC dataset. The zFAM repository is a z/OS-based File Access Manager used to store text or binary contents. The codebase contains many of the characteristics of real-world ALC codebases including complexity and program size. As such, it was selected as a representative sample of ALC code for our experimentation. In total, the ALC corpus contains 13,344 lines of code and 7,097 developer comments across 12 files.

\subsection{Evaluating the Prompting Strategy}

We employed a diverse set of metrics to measure and/or predict aspects the generated comments for various features.  For this study we took human-evaluation scores as a proxy for ground truth comment quality.  Secondarily, we consider several families of metrics that can be computed automatically---quantifying code complexity, assessing the quality of generated comments, and calculating the cost of comment generation for each model.

\subsubsection{Human Evaluation}

To measure the quality of the LLM generated output, we enlisted SMEs to evaluate both the comments produced by each LLM and the original (human-written) ground truth comments. For the MUMPS-language corpus, we had four human reviewers.  All four reviewers had a professional background in healthcare IT and a combined 16+ years of experience with the MUMPS language .  For the ALC corpus, we enlisted five reviewers, all of whom have worked extensively with IBM Z-series mainframes and legacy code (some with decades of experience). We used Label Studio \cite{LabelStudio} to present these SMEs with the rubric show in Table \ref{tab:rubric} to grade the LLM output using a 4-point scale across four categories: hallucination, readability, completeness, and usefulness.

Of the 235 developer comments in the MUMPS dataset, 177 unique source comments were evaluated by the SMEs. Of the 893 LLM-generated comments produced, 742 unique generated comments were evaluated by the SMEs. In total, the SMEs produced a total of 3,093 evaluations for the MUMPS dataset. For the ALC dataset, 127 of the 7,097 developer comments and 544 of the 21,698 LLM-generated comments were evaluated by the SMEs. In total, the SMEs produced a total of 2,268 evaluations for the ALC dataset. Across the five comment sources (four LLMs and one set of ground-truth comments), an approximately equal number of comments were reviewed from each source.\footnote{The number of comments evaluated by each reviewer were not exactly equal, because some reviewers did not complete their review assignments and reviewed slightly fewer or more comments for some sources.} For MUMPS, 2,392 comment reviews (about 77\%) overlapped with all four reviewers, and only 43 (about 1\%) were reviewed by one reviewer ---providing a set of common ratings that we use to calculate inter-rater reliability.  On ALC, 600 comment reviews (about 26\%) were rated by all five reviewers and only 143 (about 6\%) by one reviewer.



\begin{table*}[t]
\begin{center}
\begin{tabular}{|l|p{5cm}|l|}
\hline
Metric & Description & Grade Scale \\ \hline
Hallucination & Does the comment provide true information? & \begin{tabular}[c]{l}4: The comment provides only true information\\ 3: The comment provides mostly true information\\ 2: The comment provides mostly untrue information\\ 1: The comment is completely untrue\end{tabular} \\ \hline
Readability & Is the comment clear to read? & \begin{tabular}[c]{l}4: The comment is well-written\\ 3: The comment has few problems\\ 2: The comment has many problems\\ 1: The comment is unreadable\end{tabular} \\ \hline
Completeness & Does the comment address all capabilities of the relevant source code? & \begin{tabular}[c]{l}4: All essential functionality is documented\\ 3: Most essential functionality is documented\\ 2: Little essential functionality is documented\\ 1: No essential functionality is documented\end{tabular} \\ \hline
Usefulness & Is the comment useful? & \begin{tabular}[c]{l}4: The comment helps an expert programmer understand the code better\\ 3: The comment helps an average programmer understand the code better\\ 2: The comment documents only trivial functionality\\ 1: The comment is not useful at any level\end{tabular} \\ \hline
\end{tabular}
\end{center}
\caption{The rubric created for human evaluation of LLM generated output.}
\label{tab:rubric}
\end{table*}

\subsubsection{Quantifying Code Complexity }
We used the common complexity metrics of cyclomatic complexity, Halstead complexity measures, and maintainability index to quantify code complexity. Cyclomatic Complexity is a quantitative measure of the complexity of the structure and flow of control through the program, while Halstead complexity measures are a series of metrics related to the difficulty of the program to write or understand including program length, volume, difficulty, and effort. The maintainability index is calculated from the cyclomatic complexity and Halstead complexity measures and is inversely related to complexity. Therefore, more complex code, as determined by the complexity metrics, will result in a lower maintainability index while less complex code will result in a higher maintainability index. 

In addition to common complexity metrics, we used the prevalence of pain points to indicate complexity for the MUMPS dataset. Pain points are the aspects or features of the coding language that make understanding the code or following the flow of control challenging including syntax and the language's ecosystem. We measured the prevalence of six MUMPS pain points: indirection, kill/goto, terse lines, local variables, variables matching the names of built-in functions, and mathematical expressions. Indirection and goto commands make flow of control difficult to follow while kill commands, local variables, global variables, and the use of reserve words as variables make variable tracking challenging. For this experiment, we created a custom parser to measure the prevalence of pain points. The pain points were normalized by lines of code and the metrics were averaged to create a pain point metric for each file sampled from the MUMPS dataset. 


\subsubsection{Assessing Quality of Generated Comments}

To measure the quality of the LLM generated output, we used popular reference-based metrics ROUGE \cite{lin2004rouge}, $\mathrm{CHR}F$ \cite{popovic2015chrf}, and BLEU \cite{papineni2002bleu} as well as a cosine similarity score. Reference-based metrics calculate the similarity between the LLM output and defined references to measure the accuracy of the output. For this experiment, reference-based metrics compared the LLM-generated comment to the original source comment for both the MUMPS and ALC datasets. For the cosine similarity, we used OpenAI's \texttt{text-embedding-3-small} model to convert the generated and reference texts to 1,536-dimensional embeddings vectors $H(o)$ and $H(r)$, respectively, before taking the cosine of the angle between them.

In addition to evaluating the accuracy of the generated comments using reference-based metrics, the comments must also be readable to effectively support modernization efforts. To measure the readability of LLM output, we used Flesch Reading Ease and Gunning-Fog metrics. These metrics assess the readability of the output based on the number and complexity of the words in the text. 

\begin{table*}
    \begin{center}
    \resizebox{\textwidth}{!}{
        \def\arraystretch{3.0}
        \begin{tabular}{c|l|c|l}
            \hline
            & \textbf{Metric} & \textbf{Expression} & \textbf{Variables} \\
            \hline
             \multirow{5}{*}{\rotatebox{20}{\textbf{Code Complexity}}} & Cyclomatic complexity & $M = E - N + 2P$ & $E$ = edges, $N$ = nodes, $P$ = connected components \\
             & Halstead Difficulty & $\displaystyle D = \frac{\eta_1}{2}\frac{N_2}{\eta_2}$ & $\eta_1$ = distinct operators, $\eta_2$ = distinct operands, $N_2$ = total operands\\
             & Halstead Effort & $E = D \times V$ & $D$ = difficulty, $V$ = volume \\
              &Halstead Volume & $V = (N_1 + N_2)\log_2(\eta_1 + \eta_2)$ & $N_1$ = total operators \\
             & Maintainability & $M = 171 - 5.2 * \ln(V) - 0.23 * (M) - 16.2 * \ln(L)$ & $V$ = Halstead volume, $M$ = cyclomatic complexity, $L$ = total lines of code \\
             \hline
             \multirow{2}{*}{\rotatebox{20}{\textbf{Automated Quality}}} & Flesch & $\displaystyle 206.835 - 1.015 \left( \frac{W}{S} \right) - 84.6 \left( \frac{Y}{W} \right)$
     & $W$ = total words, $S$ = total sentences, $Y$ = total syllables\\
             & Gunning Fog & $\displaystyle0.4\left[\left(\frac{W}{S}\right) + 100\left(\frac{C}{W}\right)\right]$ & $C$ = complex words\\
             \hline
             \multirow{4}{*}{\rotatebox{20}{\textbf{Reference-based Quality}}} & ROUGE-N & $\displaystyle\frac{\displaystyle\sum_{S \in R}\sum_{g \in S}Count_{match}(g)}{\displaystyle\sum_{S \in R}\sum_{g \in S}Count(g)}$ & $R$ = reference texts, $S$ = sentence, $g$ = an n-gram in $S$ \\
             & BLEU & $\displaystyle\min\left(1, \frac{|O|}{|R|}\right) \prod_{i=1}^4 p_n^{w_n}$ & $O$ = output text, $R$ = reference text, $p_n^{w_n}$ = weighted $n$-gram precision \\
             & $\mathrm{CHR}F$ & $\displaystyle\mathrm{CHR}F = \frac{(1-\beta^2)\mathrm{CHR}P \cdot \mathrm{CHR}R}{\beta^2\mathrm{CHR}P + \mathrm{CHR}R}$ & $\beta^2$ = weight , $\mathrm{CHR}P$ = character precision. $\mathrm{CHR}R$ = character recall\\
             & Similarity & $\displaystyle\cos(\theta) = \frac{\displaystyle H(o)\cdot H(r)}{\displaystyle|H(o)|| H(r)|}$ & $H(o)$ = embedding of the output, $H(r)$ = embedding of the reference text\\
             \hline
        \end{tabular}
    }
    \end{center}
    \caption{Automated code complexity metrics, automated comment quality metrics, and and reference-based comment quality metrics used in this study.}
    \label{tab:complexity_formula}
\end{table*}

\subsubsection{Calculating Cost of Comment Generation}

In addition to the quality of generated comments, the time and monetary costs of comment generation are key factors to consider in the utilization of LLMs in code modernization efforts. During the execution of the comment generation task, we captured cost metrics for each LLM including monetary cost (which is proportional to the number of input and output tokens processed, each of which has their own price set by cloud providers), number of retries, processing time (which includes processing spent on retries), and number of failed generations. These metrics convey the cost and time expenditure required for each LLM to generate comments for the datasets.

\subsection{Hypotheses}

We investigated three hypotheses to determine the ability of automated measures to accurately determine the quality of LLM-generated comments for legacy software languages, namely MUMPS and ALC. Our first hypothesis evaluates our prompting strategy (describd below) in terms of a human-evaluation scoring rubric.
\begin{mdframed}
\begin{hypothesis}
On legacy software, \textbf{human evaluators} will find that comments generated by LLMs are hallucination-free, complete, readable, and useful in terms of a) comparison to human-written comments and b) having high absolute scores with ratings greater than 7 out of 10 on average.
\label{hyp:human}
\end{hypothesis}
\end{mdframed}
The second two hypotheses dig into different classes of automated metrics to test their ability to predict the (human-scored) quality of LLM-generated documentation---either by examining the complexity of the codebase and generation task themselves, or the content of the produced comments.
\begin{mdframed}
\begin{hypothesis}
Automated measures of \textbf{software complexity} and the \textbf{cost} of generating comments for legacy software will \emph{not} correlate well with human evaluation of the generated documentation quality.
\label{hyp:automated}
\end{hypothesis}
\end{mdframed}
\begin{mdframed}
\begin{hypothesis}
On legacy software, automated and \textbf{reference-based quality metrics} for generated documentation such as reading level, $\mathrm{CHR}F$, BLEU, and ROUGE will \emph{not} correlate with human evaluations of the documentation quality.
\label{hyp:reference}
\end{hypothesis}
\end{mdframed}

\section{Results}

We present the results of our human evaluation of LLMs ability to perform line-wise comment generation on legacy software, and then examine the ability of automated and reference-based metrics to predict that performance.

\subsection{Human Evaluation of Comments}

On the MUMPS benchmark, we observed moderate to good \cite{koo2016guideline} inter-rater reliability among human evaluators, as measured by intraclass correlation (ICC) ($ICC(2, k)$=0.73 [0.63, 0.8] for hallucination, 0.65 [0.6, 0.69] for readability, 0.89 [0.87, 0.91] for completeness, and 0.85 [0.81, 0.88] for usefulness, where square brackets indicate 95\% confidence intervals).
On the ALC benchmark, however, we see poor inter-rater reliability ($ICC(2, k)$=0.22 [-0.1, 0.48] for hallucination, 0.16 [0, 0.31] for readability, 0.21 [-0.15, 0.48] for completeness, and 0.12 [0.17, 0.37] for usefulness). These results indicate differences in opinion among reviewers and highlight a challenge for modernization: comment quality is difficult to measure and even highly experience experts, like our reviewers, may have varying ideas on the qualities that make a useful/readable/etc.\ comment. 

\begin{figure*}
    \centering
    \includegraphics[width=0.49\textwidth]{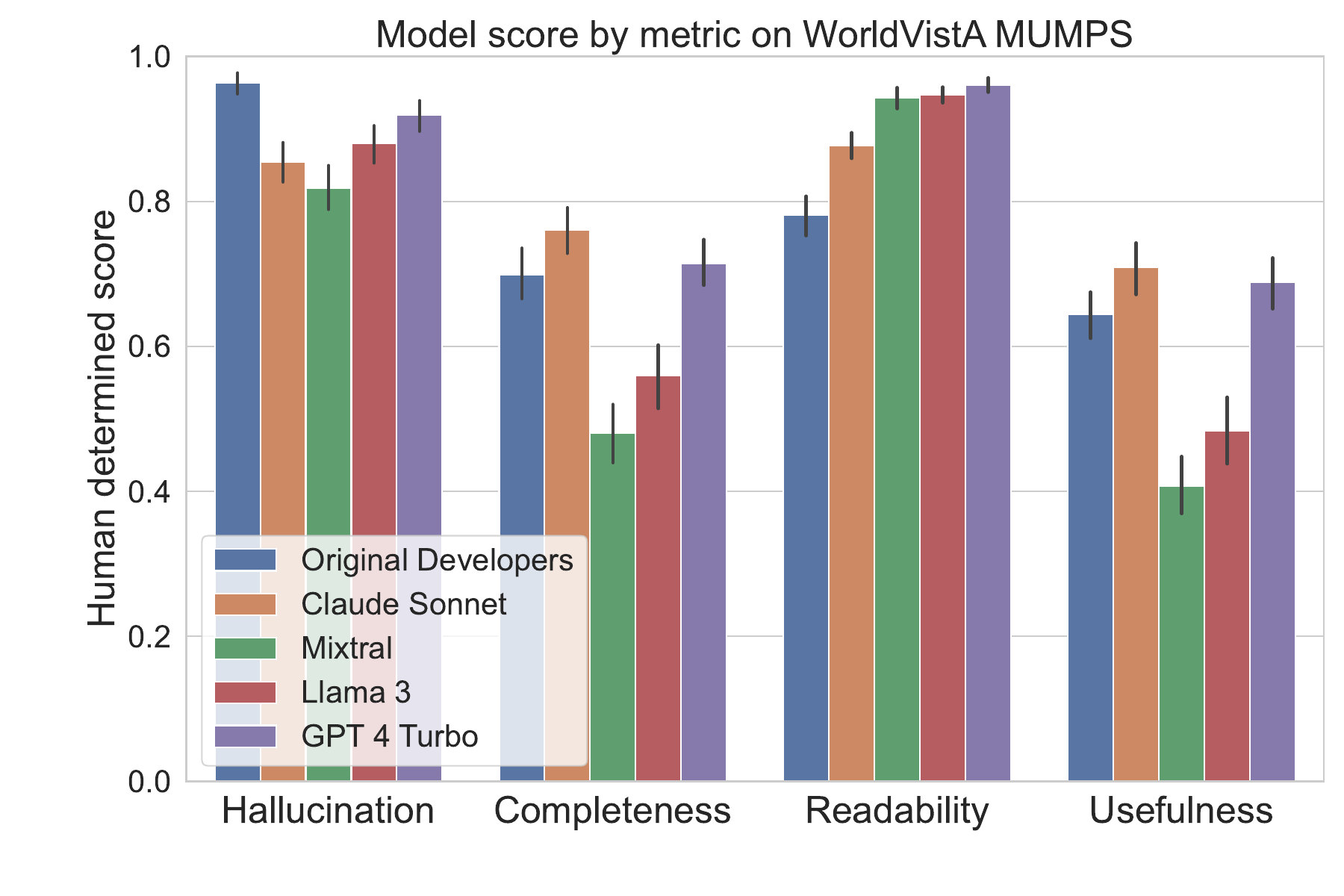}
    \includegraphics[width=0.49\textwidth]{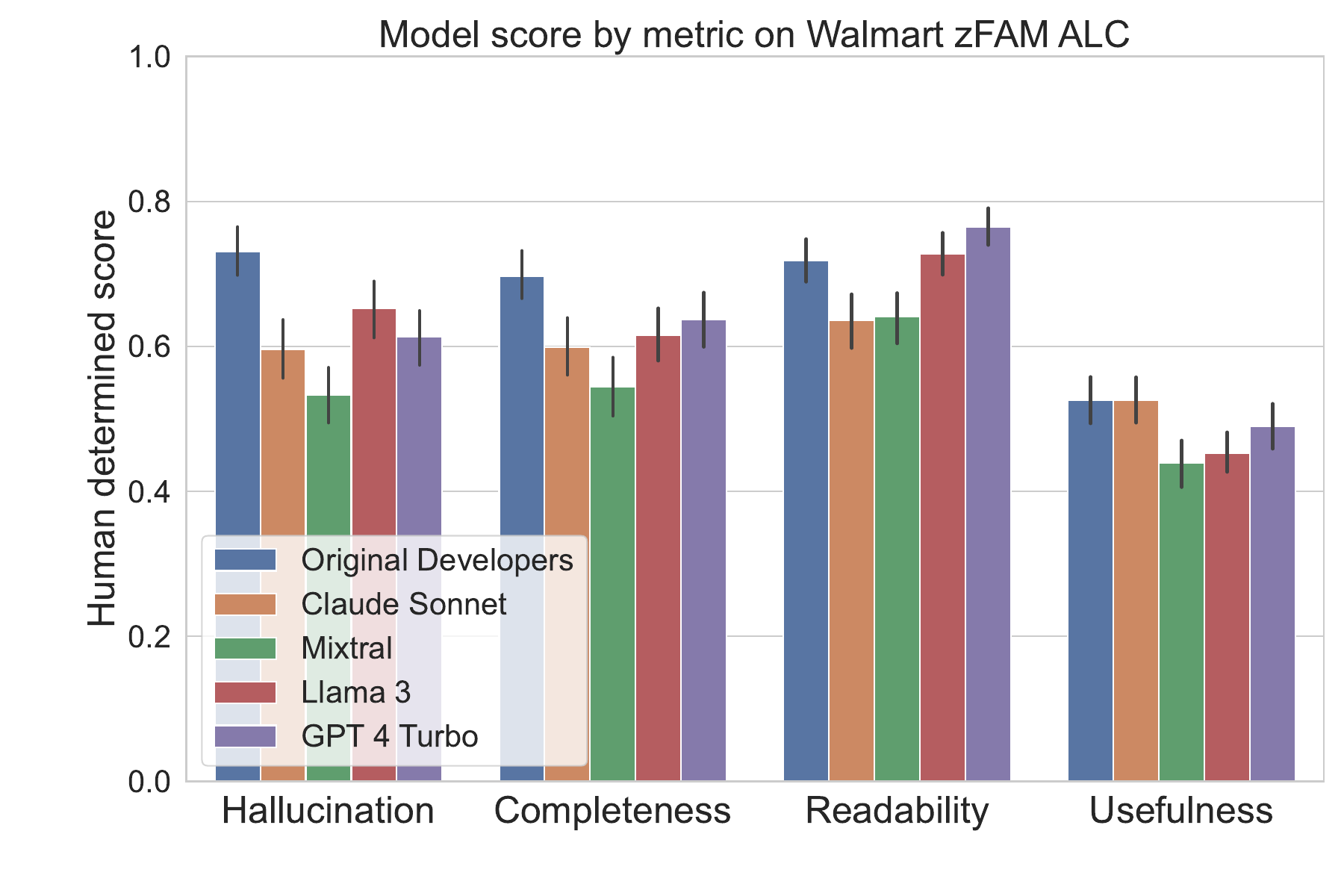}
    \caption{Mean human evaluation scores for code comments generated by different LLMs (with 95\% confidence intervals on the mean), for the MUMPS (WorldVistA) dataset on the left and ALC (zFAM) datasets on the right.}
    \label{fig:human_eval_bars}
\end{figure*}

The evaluation results for the ground truth and all four models are shown in Figure~\ref{fig:human_eval_bars}.  In most cases, comments generated for the MUMPS dataset are rated considerably higher on average by SMEs than comments generated for ALC. Compared to the original (human-written) ground-truth comments (shown in the blue, leftmost bar within each group), our SMEs rated comments generated by the four LLMs relatively favorably across the four scales.  Only a few large performance gaps occur---Mixtral and Llama3 fair poorly on generating useful and complete MUMPS code, for instance, while the other two models perform similarly to the ground truth. We find that human reviewers are especially conservative in judging the usefulness of comments.  Readability is an area where LLM-generated comments are strong, particularly on MUMPS, where all models outperform ground truth by a noteworthy margin.  This is perhaps not surprising when we consider that real-world comments written by humans are often sparse and obscure. On both datasets, we observe a measurable tendency for the LLMs to hallucinate (i.e.\ to generate factually incorrect comments).  GPT-4 Turbo nonetheless achieves an average rating of 9.1 out of 10 for factualness on the MUMPS code, and Llama3 comes close to the factualness level of ground-truth comments on ALC (6.53/10 vs 7.31/10, respectively). 

Comparison to ground truth is an imperfect metric for source code, however, because each dataset's original corpus of human-written comments are not necessarily of good quality.  In terms of absolute scores, the LLMs perform well overall (that is, achieving greater than 8 out of 10) on creating readable and hallucination-free documentation for MUMPS, and moderately well (greater than 7 out of 10 for Claude and GPT 4 Turbo) at providing complete documentation.  Only Claude Sonnet achieved greater than 7 out of 10 on usefulness ratings, however. On our ALC dataset the picture is different: only on readability do we see any LLM-generated comments rated above 7/10 (Llama3 and GPT-4 Turbo at 7.28 and 7.65, respectively).

In summary, our SME reviewers found that LLMs create comments that are highly readable and factual on MUMPS, and that generally rate similarly to human-written ground truth comments in both languages---but that the ALC comments are low quality overall. We find \textbf{Hypothesis~\ref{hyp:human}(a) to be confirmed} for both ALC and MUMPS across the four LLMs used in the experimentation, human evaluation finds that comments generated by LLMs tend to be nearly as hallucination-free, complete, readable, and (with some exceptions) as useful as human-written comments.  But we find \textbf{Hypothesis~\ref{hyp:human}(b) to be confirmed only for MUMPS and \emph{not} for ALC}.By absolute scores, the LLM-generated comments achieve poor ratings almost across the board on ALC. This performance disparity may be attributed to the complexity of the ALC language, further exacerbated by the challenges SMEs faced in defining what constitutes a high-quality ALC comment. ALC diverges structurally and syntactically from contemporary programming languages, whereas MUMPS shows a closer alignment with modern language constructs. This highlights the potential of LLM generated documentation for legacy software modernization efforts as well as the challenges surrounding evaluation to determine the quality of generated documentation at scale. 

We do not observe a relationship between context-window size and the performance of the four models under study. While our chunking strategy provides input of various sizes in accordance with the varibale context-window sizes of the models (ex.\ 32k for Mixtral versus 200k for Claude Sonnet), this additional information appears to have had little impact on the human-rated quality of resulting line-wise comments. We anticipate that context will play a more critical role in higher-level software documentation (such as code summarization for full subroutines and modules), where being able to describe relationships between distributed pieces of code is more important.

\subsection{Complexity and Cost Metrics}

Human evaluation alone is costly and, as such, is insufficient to adapt LLM-based comment generation techniques to new real-world applications.  If automated or semi-automated metrics are able to assist in verifying a system's performance in new settings, it would go a long way to making these tools easier to adopt.

We find, however, that the various complexity and cost metrics that we present in this study---while easy to compute---correlate poorly with human evaluations of comment quality.  Table~\ref{tab:complexity_metrics} shows Pearson correlation coefficients for these metrics on both datasets.  Many metrics show no statistically significant\footnote{As we were only interested in a general view of the potential of these metrics to correlate with comment quality, we did not perform a Bonferroni correction on these results to control for the inflated risk of Type-II errors.  Strong conclusions should not be drawn here, then, about individual correlation values.} ($p < 0.05$) correlation with hallucination, completeness, readability, or usefulness. The absolute value of the correlations that are significant are all beneath 0.25---except for the inverse correlation between processing time and usefulness, which has a correlation of -0.33. Interestingly, this suggests that the longer an LLM spends generating a comment, the less useful it is (up to a weak correlation). This effect seems to be independent of the number of retries, which (while retries are included in processing time) shows no significant correlation by itself.  Overall we find \textbf{Hypothesis~\ref{hyp:automated} to be confirmed} for both ALC and MUMPS across the four LLMs used in the experimentation: automated metrics of both static complexity and runtime cost correlate poorly with the quality of comments produced by LLMs.  While larger sample sizes might lead to more of the correlations we measured becoming statistically significant, we still observe low correlation strengths across the board (even in significant cases).  This may partly be due to the variability across SME opinions that we observed in the ICC figures above for ALC---but it also likely reflects the deep difficulty of understanding and summarizing legacy assembly instructions in ways that are useful to humans.


\begin{table*}
\begin{center}
\begin{tabular}{|l|l|cccc|}
\hline
\textbf{Dataset} & \textbf{Metric} & \textbf{Hallucination} & \textbf{Completeness} & \textbf{Readability} & \textbf{Usefulness} \\
\hline
\multirow{10}{*}{Walmart ALC} & \textbf{Cost} & 0.06 & 0.09 & \textbf{0.21} & 0.06 \\
& \textbf{Cyclomatic Complexity} & -0.01 & -0.02 & -0.02 & 0.05 \\
& \textbf{Difficulty} & -0.02 & -0.01 & 0.05 & 0.04 \\
& \textbf{Effort} & -0.01 & -0.02 & -0.01 & 0.05 \\
& \textbf{Maintainability} & -0.02 & 0.00 & 0.07 & 0.00 \\
& \textbf{Processing Time} & \textbf{-0.11} & -0.03 & -0.09 & -0.05 \\
& \textbf{Retries} & 0.07 & 0.02 & 0.06 & -0.07 \\
& \textbf{Volume} & -0.01 & -0.02 & -0.01 & 0.05 \\
\hline
\multirow{14}{*}{MUMPS} & \textbf{Cost} & 0.02 & -0.01 & 0.03 & 0.01 \\
& \textbf{Cyclomatic Complexity} & \textbf{-0.10*} & \textbf{-0.16*} & \textbf{-0.24*} & \textbf{-0.09*} \\
& \textbf{Difficulty} & \textbf{-0.09*} & \textbf{-0.22*} & \textbf{-0.20*} & \textbf{-0.16*} \\
& \textbf{Effort} & -0.08 & \textbf{-0.21*} & \textbf{-0.19*} & \textbf{-0.16*} \\
& \textbf{Maintainability} & -0.09 & \textbf{-0.15*} & \textbf{-0.21*} & \textbf{-0.09*} \\
& \textbf{Pain Point Indirection} & -0.07 & \textbf{-0.23*} & \textbf{-0.15*} & \textbf{-0.17*} \\
& \textbf{Pain Point Kill Goto} & -0.07 & \textbf{-0.25*} & \textbf{-0.13*} & \textbf{-0.18*} \\
& \textbf{Pain Point Lines of Code} & -0.08 & \textbf{-0.18*} & \textbf{-0.17*} & \textbf{-0.13*} \\
& \textbf{Pain Point Locals} & \textbf{-0.09*} & \textbf{-0.21*} & \textbf{-0.18*} & \textbf{-0.15*} \\
& \textbf{Pain Point Math} & -0.06 & \textbf{-0.10*} & -0.04 & \textbf{-0.10*} \\
& \textbf{Pain Point Terse} & -0.08 & \textbf{-0.19*} & \textbf{-0.17*} & \textbf{-0.12*} \\
& \textbf{Pain Point Variables Matching Functions} & \textbf{-0.09*} & 0.02 & 0.02 & 0.00\\
& \textbf{Processing Time} & \textbf{-0.11*} & \textbf{-0.16*} & -0.04 & -0.13* \\
\hline
\end{tabular}
\end{center}
\label{tab:complexity_metrics}
\caption{Correlations between automated complexity, pain-point, and runtime cost metrics and human-evaluation scores on the Walmart ALC (top) and MUMPS (bottom) datasets.  Negative values indicate that higher values of the metric are associated with lower scores, and values further from zero indicate stronger relationships. Bolded values with asterisks indicate statistically significant correlations.}
\end{table*}

\subsection{Quality Metrics}

Reference-based metrics occupy a middle ground between cheap and easy-to-compute automated measures and the full expense of 100\% human evaluation.  In practice, organizations may use a small corpus of human-generated ground truth along with reference-based metrics to evaluate LLM-based comment generation tools on their codebase.  Once the ground truth is created, these metrics can be run automatically against any number of new models or hyperparameter configurations---allowing accelerated research into ways to improve comment generation.

Unfortunately, we find that neither fully automated quality metrics nor reference-based metrics correlate well with human evaluations of quality.  As shown in Table~\ref{tab:reference_based_results}, of the statistically significant correlations, only a handful achieve a Pearson correlation of greater than 0.25. The highest correlation is between the cosine similarity score and usefulness---at a correlation of just 0.34.  Overall this similarity metric has the most robust performance aross the four human-evaluation categories, and thus is the most interesting prospect going forward for a reference-based quality metric. We observe that comments that have higher complexity in terms of Flesch and Gunning-Fog readability measures tend to be a little bit more complete (0.17 and 0.20, respectively)---but these comments tend to exhibit a similar increase in hallucinations (0.24 and 0.26). Overall, because all of these correlations are small in magnitude, we find \textbf{Hypothesis~\ref{hyp:reference} to be confirmed} for both ALC and MUMPS across the four LLMs used in the experimentation: automated and reference-based measures of generated comment quality are poor predictors of their true quality as rated by human subject-matter experts.


\begin{table*}
\begin{center}
\begin{tabular}{|l|l|cccc|}
\hline
\textbf{Dataset} & \textbf{Metric} & \textbf{Hallucination} & \textbf{Completeness} & \textbf{Readability} & \textbf{Usefulness} \\
\hline
\multirow{4}{*}{Walmart ALC} & \textbf{BLEU} & 0.13 & \textbf{0.15*} & 0.03 & 0.02 \\
& \textbf{Flesch} & \textbf{0.24*} & \textbf{0.17*} & -0.13 & -0.01 \\
& \textbf{Gunning Fog} & \textbf{0.26*} & \textbf{0.20*} & -0.11 & 0.01 \\
& \textbf{CHRF} & \textbf{0.24*} & \textbf{0.27*} & 0.07 & 0.11 \\
& \textbf{ROUGE} & 0.11 & 0.10 & 0.02 & -0.04 \\
& \textbf{Similarity Score} & \textbf{0.31*} & \textbf{0.33*} & 0.10 & \textbf{0.23*} \\
\hline
\multirow{4}{*}{MUMPS} & \textbf{BLEU} & 0.03 & \textbf{0.11*} & 0.05 & \textbf{0.09*} \\
& \textbf{Flesch} & -0.01 & \textbf{-0.10*} & \textbf{-0.10*} & \textbf{-0.06*} \\
& \textbf{Gunning Fog} & -0.02 & \textbf{-0.09*} & \textbf{-0.11*} & -0.06 \\
& \textbf{CHRF} & \textbf{0.06*} & \textbf{0.23*} & \textbf{0.09*} & \textbf{0.22*} \\
& \textbf{ROUGE} & \textbf{0.09*} & \textbf{0.14*} & 0.06 & \textbf{0.16*} \\
& \textbf{Similarity Score} & \textbf{0.19*} & \textbf{0.32*} & \textbf{0.13*} & \textbf{0.34*} \\
\hline
\end{tabular}
\end{center}
\caption{Correlations between automated and reference-based LLM quality metrics and human-evaluation scores on the Walmart ALC (top) and MUMPS (bottom) datasets. Negative values indicate that higher values of the metric are associated with lower scores, and values further from zero indicate stronger relationships. Bolded values with asterisks indicate statistically significant correlations.}
\label{tab:reference_based_results}
\end{table*}

\section{Discussion}

Legacy code modernization is an exceedingly demanding task that has a reputation for being difficult for any organization to achieve successfully (let alone in-budget!). The market contains many vendors promising solutions that can accelerate modernization, and the growing body of evidence around LLM-based software understanding suggests that rapid progress can be expected over the next few years. But, based on our experience with real-world IT decision makers and legacy software SMEs, we recognize that there is a growing need for methods of evaluating whether or not new tools pose opportunities to automate and accelerate modernization efforts.

Documentation generation---and more specifically the generation of line-wise comments that we study here---plays one small part in the broader modernization lifecycle (which itself may involve many decisions about when to translate, refactor, rewrite, or re-deploy and containerize various legacy components).  What kind of evidence do organizations and developers need in order to justify placing their trust and resources in a modernization workflow that includes generative LLM components for tasks such as automated documentation?

When it comes to the practical use of comment-generation tools, developers are waiting to see how the technology matures. Research has often cited a lack of good benchmark datasets as a challenge in the documentation-generation field \cite{leclair2019recommendations}. Additionally, there is a tendency for model performance to vary greatly across different datasets---such that performance results on one benchmark are likely to generalize poorly to new codebases \cite{stapleton2020human}. The popular benchmarks that do exist are focused exclusively on mainstream languages (especially Java)---with, again, a gap in the literature when it comes to their performance on legacy languages and codebases. In a wide-ranging study of practitioner perceptions of automated comment generation, Hu et al. find that software developers cite concerns that such tools will be limited to restating ``what the code does,'' and will perform poorly at the most valuable aspects of documentation---such as describing \emph{why} the code works the way it does, how it can be used, and how it connects to other aspects of the system
\cite{hu2022practitioners}. They also find that practitioners are especially skeptical of n-gram-based performance metrics for documentation, such as BLEU, ROUGE, and $\mathrm{CHR}F$ \cite{papineni2002bleu,lin2004rouge,popovic2015chrf}. Reference-based metrics like these originate in the machine translation and text summarization community \cite{li2024leveraging} as a means of quantifying similarity to human-written ground truth---but similarity alone does not seem sufficient to quantify documentation quality.

\subsection{Conclusions}

This paper evaluated the performance of LLMs for code understanding (specifically line-level code summarization) on legacy software.  Broadly, we find reason to be optimistic. Even though MUMPS and mainframe assembly are aging languages with small, niche communities and relatively little publicly available code for foundation models to be trained on, we found that mainstream LLMs generate comments of comparable quality to ones that are written by human developers (Hypothesis 1(a)). We found that LLM generation of MUMPS comments performs rather well across the board, but that writing comments for ALC is challenging both for humans and machines (Hypothesis 1(b)). We speculate that the performance difference between ALC and MUMPS arises because MUMPS' structure and flow are more similar to modern languages, whereas ALC's structure and flow differ significantly from them.

The question that practitioners have, however, is how these automation techniques can be expected to perform on their proprietary code.  No set of academic benchmarks can fully answer this question, and organizations will need to conduct some degree of experimentation or measurement on their own systems to test out the feasibility and validate the results of automation.  Turning our attention to code complexity metrics and LLM runtime cost metrics---which can be measured fully automatically with no ground truth comments to compare against---we found that none of them correlate strongly with any of the four dimensions of our human evaluation of comment quality (Hypothesis 2).  These correlations were especially weak on ALC code.  Looking further at direct measures of comment quality like reading level, BLEU, and ROUGE (among others), we again found no strong correlations with human assessments of comments (Hypothesis 3)---though here we do see a number of metrics that correlate significantly even on ALC code.

\subsection{Future Work}

The comment generation method we tested here using out-of-the-box foundation models was effective at solving the initial problems of incomplete and mutated output, but there are many avenues that could be taken to improve comment quality.  Industry players are already demonstrating the benefits of domain-specific fine tuning for code modernization tasks (ex.\ for COBOL \cite{IBM2023}), and we would like to investigate how (parameter-efficient) fine tuning can improve performance on especially difficult instances of code modernization \cite{han2024parameter}.  Getting access to realistic legacy code to fine tune on is a barrier in the field, however, as open-source repositories exhibiting realistically complex code in legacy languages are not common.  Continuing to test methods of this kind on a broader set of legacy code benchmark is thus vitally important for the field, especially since there is strong evidence that performance of ML models on code understanding tasks varies greatly across codebases and the results we present here do not guarantee similar results on specific legacy systems \cite{stapleton2020human}.

Another significant set of design decisions we made here surround our greedy chunking strategy for feeding code into LLMs with different context window sizes.  Little evidence has been published around what kind of chunking strategies work best for source-code-related tasks. We expect to study alternative chunking strategies in our future work.


On the metrics side, our results reiterate an \emph{alignment problem} that has been a growing concern in the literature the past few years \cite{hu2022correlating,shi2022evaluation,stapleton2020human}: we do not have good automated or semi-automated metrics for evaluating the quality of LLM-generated documentation.  To date, the only way to provide confidence in the correctness of generated documentation is to have a human expert spend great amounts of time reading and verifying large amounts of the system's output.  A few efforts have begun to examine alternate metrics to close this gap, perhaps based on machine learning \cite{haque2022semantic,mastropaolo2024evaluating}, but much further investigation is needed especially in the IT modernization sphere, where legacy systems pose unique challenges that are not well-represented in open-source benchmarks.

\section*{Acknowledgment}

\copyright The MITRE Corporation. All Rights Reserved. Approved for Public Release; Distribution Unlimited. Public Release Case Number 24-3245.

\bibliographystyle{IEEEtran}
\bibliography{main}

\end{document}